\documentclass{ecai}
\usepackage{graphicx}
\usepackage{latexsym}

\usepackage{booktabs}
\usepackage{algorithm}
\usepackage{algorithmic}
\usepackage{paralist}
\usepackage{multirow}
\usepackage[switch]{lineno}
\usepackage{color}
\usepackage{xcolor,soul}

\newcommand{\hlg}[2][green]{{\sethlcolor{#1}\hl{#2}}}
\newcommand{\hlr}[2][orange]{{\sethlcolor{#1}\hl{#2}}}

\begin{document}

\begin{frontmatter}

\title{Rationale-Guided Few-Shot Classification\\ to Detect Abusive Language}

\author{\fnms{Punyajoy}~\snm{Saha}\thanks{Corresponding Author. Email: punyajoys@iitkgp.ac.in}}
\author{\fnms{Divyanshu}~\snm{Sheth}}
\author{\fnms{Kushal}~\snm{Kedia}}
\author{\fnms{Binny}~\snm{Mathew}}
\author{\fnms{Animesh}~\snm{Mukherjee}}


\address{Indian Institute of Technology, Kharagpur}

\begin{abstract}
Abusive language is a concerning problem in online social media. Past research on detecting abusive language covers different platforms, languages, demographies, etc. However, models trained using these datasets do not perform well in cross-domain evaluation settings. To overcome this, a common strategy is to use a few samples from the target domain to train models to get better performance in that domain (cross-domain few-shot training). However, this might cause the models to overfit the artefacts of those samples. A compelling solution could be to guide the models toward rationales, i.e., spans of text that justify the text's label. This method has been found to improve model performance in the in-domain setting across various NLP tasks. In this paper, we propose RGFS (Rationale-Guided Few-Shot Classification) for abusive language detection. We first build a multitask learning setup to jointly learn rationales, targets, and labels, and find a significant improvement of 6\% macro F1 on the rationale detection task over training solely rationale classifiers. We introduce two rationale-integrated BERT-based architectures (the RGFS models) and evaluate our systems over five different abusive language datasets, finding that in the few-shot classification setting, RGFS-based models outperform baseline models by about 7\% in macro F1 scores and perform competitively to models finetuned on other source domains. Furthermore, RGFS-based models outperform LIME/SHAP-based approaches in terms of plausibility and are close in performance in terms of faithfulness. \\
  \noindent\textbf{Disclaimer:} This paper contains material that many will find offensive or hateful. However, this cannot be avoided owing to the nature of the work.
\end{abstract}

\end{frontmatter}
\section{Introduction}

Abusive language has become a perpetual problem in today's online social media. An ever-increasing number of individuals are falling prey to online harassment, abuse and cyber-bullying as established in a recent study by Pew Research~\cite{TheState71:online}. In the online setting, such abusive behaviours can lead to traumatization of the victims~\cite{vedeler2019hate}, affecting them psychologically. Furthermore, widespread usage of such content may lead to increased bias against the target community, making violence normative~\cite{Dehumani67:online}. Many gruesome incidents like the mass shooting at Pittsburgh synagogue\footnote{\url{https://en.wikipedia.org/wiki/Pittsburgh_synagogue_shooting}}, the Charlottesville car attack\footnote{\url{https://en.wikipedia.org/wiki/Charlottesville_car_attack}}, etc. are all caused by perpetrators consuming/producing such abusive content. 

In response, various social media companies have implemented policies to moderate the content on their platforms\footnote{\url{https://help.twitter.com/en/rules-and-policies/abusive-behavior}(Accessed on 05/02/2021)}. These are primarily handled by moderators who manually delete posts that violate community guidelines specific to the platform~\cite{gillespie2018custodians}. The main issue with such a system is the sheer volume of content to be reviewed, which in many cases does not leave enough time for the moderator to arrive at a decision\footnote{\url{https://tinyurl.com/2m4jdw54}}. Furthermore, many moderators complain about psychological effects caused due to moderation of such abusive content. To help the moderators, various social media platforms are trying to proactively filter such abusive content using recent NLP architectures like transformers\footnote{\url{https://ai.facebook.com/blog/how-ai-is-getting-better-at-detecting-hate-speech/}}. Such filtering techniques also need human support, in case of complex examples, but the manual effort of moderators should reduce a lot. This will benefit the moderation system as a whole.

However, reliably detecting abusive language is a challenging problem. Efforts using lexicon-based systems~\cite{elsherief2018hate} and machine learning~\cite{davidson} have been made to detect abusive speech in online social media. As highlighted by Mishra~\cite{mishra2019tackling}, one of the issues is \textit{domain shift}, i.e., when the training and test data come from different distributions. This domain shift is common in abusive language research due to the variability in annotation, demography and topics~\cite{vidgen2020directions}. This calls for models/pipelines which can effectively domain transfer across different datasets with zero or few annotated datapoints. Though there has been a considerable volume of research in cross-domain transfer in other NLP problems like machine translation~\cite{chu2020survey}, the ever-changing nature of abusive language makes it a very crucial problem~\cite{10.1145/3415163}.

Another important aspect missing in the abuse detection pipeline is explainability~\cite{mishra2021user}. A lot of the current research in NLP concentrates on not just detections but providing explanations behind the detections as well~\cite{danilevsky2020survey}. A subset of such research has been involved in creating datasets that contain the annotators' reasoning in some form, i.e., the spans of text or rationales, textual reasons, etc.~\cite{deyoung2020eraser}. We hypothesize that the nucleus of abusiveness lies in certain text spans, i.e., rationales in typical hate posts. Hence, it is essential to give additional attention to these spans compared to the whole post. This attention can also be provided by including novel architectural changes in the computational model. As highlighted in \cite{mathew2021hatexplain}, such rationales, when used as a feedback to the model, can also help in improving abuse classification and reduce the unintended bias of the model toward various target communities. 

In this work, we investigate if the use of such rationales can help better cross domain few-shot classification\footnote{For our case, cross domain few-shot classification is a variant of few-shot classification where we have a few samples from the target domain and can use some information from the source.} in various datasets.  We propose RGFS -- Rationale-Guided Few-Shot Classification to Detect Abusive Language, which uses an attention framework to introduce rationales, i.e., text spans which justify the classification labels, into the prediction pipeline. To train models to predict rationales in text, we utilise a multitask framework which jointly learns labels, rationales, and targets for a given text. We use this rationale predictor - called BERT-RLT (BERT-Rationale-Label-Target) - to predict rationales for datapoints in an unseen dataset and integrate it with RGFS, using a few labeled samples for training. To evaluate the pipeline, we use five different datasets having different numbers of labels, collected from different timelines and using slightly different annotation guidelines. We observe that:

\begin{compactitem}
    \item The multitask framework that jointly learns rationales, labels and targets outperforms the model learning only rationales by 6\% in terms of rationale classification macro F1 score.
    \item In the cross domain few-shot setting, our proposed model - RGFS outperforms BERT models by about 7\% in terms of macro F1 scores and performs comparably to models already fine-tuned on a similar dataset.
    \item Predicted rationales used in RGFS-based models are more plausible than LIME/SHAP-based rationales. The predicted rationales outperform LIME/SHAP-based rationales by around 40\% and 30\% in terms of AUPRC and token-F1 scores respectively.
    \item Further, rationale-based explanation when utilised with the RGFS-based architecture achieves comparable performance to LIME/SHAP-based explanation in terms of faithfulness.  
\end{compactitem}

\noindent Please refer to Appendix\footnote{\url{https://rb.gy/kbwqz}} for additional details about the architectures, dataset and discussion on rationales. The code is added here\footnote{\url{https://github.com/punyajoy/RGFS_ECAI}}.

\section{Related work}
\subsection{Few-shot learning}

Research in cross domain few-shot classification mostly focuses on instance selection~\cite{ruder2017learning}. In the semi-supervised setting, a few labelled as well as unlabelled datasets are used for the domain transfer~\cite{ijcai2022p398}. Some of the methods used include maximising the label entropy for the unlabelled data~\cite{saito2019semi} or continuing pre-training using some auxiliary task~\cite{dontstoppretraining2020} before training with the labelled data.

In abusive language research literature, researchers have studied cross-domain performance across different datasets and found that the percentage of positive examples~\cite{swamy-etal-2019-studying} and the in-domain performance of the classifier~\cite{FORTUNA2021102524} are correlated with cross-domain performance. \cite{Waseem2018} used a multi-task learning framework where auxiliary tasks were learnt on datasets from different distributions to improve performance on the target task, using a simple classifier.
In recent work, Fortuna~\cite{FORTUNA2021102524} show that transformer-based models like \textsc{BERT}~\cite{devlin-etal-2019-bert} are already more generalisable compared to the previous models. 

Although cross-domain performance has been extensively studied for abusive language research literature, there is a lack of evaluation in few-shot classification setups. Aluru~\cite{aluru2021deep} performed an extensive study on several multilingual datasets in a few-shot setting where the authors focused on variation of abuse detection performance with respect to languages. Also,  Stappen~\cite{stappen2020cross} introduced a new architecture `AXEL' to improve cross lingual zero-shot and few-shot classification. Their study, performed across just two datasets, did not include explainability analysis.

\subsection{Explainability / Interpretability}

In recent years, NLP research has focused more on black box techniques at the expense of less interpretable models. As a remedy, different local post-hoc techniques like LIME~\cite{lime} and TREPAN~\cite{confalonieri2019trepan} have been introduced to explain the prediction of such black box models. With the advent of transformer models like \textsc{BERT}, a section of research is also focusing on understanding BERT's inner working~\cite{tenney-etal-2020-language} by visualising its internal layers and utilising BERT's attention to generate the explanation using reinforcement learning~\cite{ijcai2022p102}. Once an explanation is generated, it is crucial to measure its reliability. One of the methods is to compare it with ground truth rationales~\cite{zaidan-etal-2007-using}. DeYoung~\cite{deyoung2020eraser} compiled previous explainable works and provided several metrics to draw comparisons with ground truth rationales. In abusive language literature, a recent study~\cite{mathew2021hatexplain} showed that rationales improve hate speech classification and also help in reducing unintended biases towards target communities.

In a recent work in the computer vision community, Schmidt~\cite{schmidt2021explainability} proposed a loss using a saliency map to help in-domain generalisation for image recognition. Furthermore, Zhang~\cite{10.1145/3437963.3441758} proposed an Explain-Then-Predict framework for document level tasks where they first extracted the rationales and then used the extracted rationales only for prediction. This is different for short sentence classification, where removing the non-rationales can result in the sentence losing its meaning. There are also some research works that show improvement in low/few-shot tasks with human rationales; however, they mostly evaluate in in-domain settings. Melamud~\cite{melamud-etal-2019-combining} study if pre-training with such rationales can help in few-shot performance. This study was further extended to include rationale-based self-training to improve few-shot performance~\cite{bhat2021self}. Both these research works focus mostly on the performance of the models in light of rationales but do not discuss the explainability of these models.

In this work, we study if rationales can help in improving cross domain few-shot classification for abusive language detection if a target dataset has a few labelled samples but does not have annotated rationales. Furthermore, while previous studies have only utilised the rationale prediction as an auxiliary task, we use rationale predictions in an attention-based framework to improve upon base models in terms of both performance and explainability.

\section{Datasets}
In this section, we describe the datasets used in our work. To train the rationale extractor, we used the \textbf{HX: HateXplain}~\cite{mathew2021hatexplain} dataset, which contains the classification label, rationales, and targets annotated per post. To provide the model with rationales as feedback, the rationales by each annotator are converted into into Boolean vectors. Values in these Boolean vectors are 1 when the corresponding token (word) in the text is a part of a rationale. To create the \textit{ground truth rationales}, each token in the text is denoted as a rationale if at least two annotators have highlighted it as a rationale. The final ground truth rationales are Boolean vectors, considering the above constraint. 

\subsection{What are rationales?}

 Rationales are sets of words or phrases in text that justify the text's classification into a label or category. We consider spans of consecutive words marked as rationales by annotators on the HateXplain (HX) dataset for our analysis. The top 5 frequent rationales seen include common slur/derogatory terms like ‘ni**er, bi**h, k**e, ass. We note that about 93\% of the 6021 unique rationale phrases have a count of 1. Examples of some rationales for classifying a text as hate speech include “evil mu*rat cult”, “can not speak properly lack basic knowledge of biology”, “jew is just a ni**er turned inside out”, etc. The average length of rationale phrases is about 6 words-- many rationales are seen to have multiple words in them.

\subsection{Datasets for evaluation}
\label{eval_datasets}
To evaluate our RGFS models, we use five popular abusive language datasets - \textbf{DA: Davidson}~\cite{davidson}, \textbf{FA: Founta}~\cite{founta2018large}, \textbf{OD: Olid}~\cite{zampieri-etal-2019-predicting}, \textbf{BA: Basile}~\cite{basile-etal-2019-semeval} and \textbf{WH: Waseem \& Hovy}~\cite{waseem-hovy-2016-hateful} and use the same labels as present in these datasets.  These datasets differ in their choice of class labels, methods of annotation, geographies \& diversities of source users in their posts, communities \& groups targeted and periods of data collection.  Details about the datasets are presented in Table \ref{tab:dataset_desc} (see Appendix for more details about the datasets).

\begin{table*}
\centering
\caption{The total dataset size and the number of datapoints available per label for each dataset being used.}
\label{tab:dataset_desc}

\begin{tabular}{cccc}
\textbf{Dataset} & \textbf{Abbv.} & \textbf{Labels(\#numbers of datapoints)} & \textbf{Size}  \\ 
\hline
HateXplain       & HX              & \textit{Hate speech} (5,935), \textit{}{Offensive} (5,480), \textit{Normal} (7,814)                                  & 19,229                \\\hline
Founta et al\.    & FA              & \textit{Hateful} (4,948), \textit{Abusive} (27,037), \textit{Normal} (53,790)                            & 85,775               \\\hline
Davidson et al\.  & DA              & \textit{Hate speech} (1,430), \textit{Offensive} (19,190),  \textit{Normal} (4,163)                               & 24,783                \\\hline
OLID             & OD              & \textit{Offensive} (4,640), \textit{Not Offensive} (9,460)                           & 14,099                \\\hline
Basile et al\.    & BA              & \textit{Hateful} (5,390), \textit{Non-hateful} (7,415)                             & 12,805                \\\hline
Waseem \& Hovy  & WH              & \textit{Racism} (13), \textit{Sexism} (2855) \textit{Normal} (8050)                                                               & 10,018                 \\
\end{tabular}
\end{table*}

\subsection{A note about the datasets} 

Biases in annotations of these datasets have been noted by earlier authors. The definition of abuse varies across different datasets \cite{Ro_Rist_Carbonell_Cabrera_Kurowsky_Wojatzki_2016} and oftentimes these definitions are incompatible \cite{fortuna-etal-2020-toxic}.
Awal \cite{awal2020analyzing} noted inconsistencies among three popular abusive language datasets: Davidson, Founta and Waseem~\cite{waseem-hovy-2016-hateful}. However, with such a subjective and difficult annotation process, such inconsistencies are unavoidable.

\section{Methodology}

\subsection{Base model}
As a base model, we use \textsc{BERT}~\cite{devlin-etal-2019-bert} pre-trained on English data.\footnote{We use the bert-base-uncased model having 12-layers, 768-hidden, 12-heads, 110M parameters.} 
We perform uniform pre-processing by normalising usernames and links in datasets. Finally, we pass the inputs through the text processing pipeline --- \textit{ekphrasis} ~\cite{baziotis-pelekis-doulkeridis:2017:SemEval2}. As a baseline for classification, we attach a classifier layer on top of BERT (henceforth, BERT-L model).

\subsection{Rationale extraction}

To detect rationales, we add a token classifier layer~\cite{devlin-etal-2019-bert} to classify each token based on whether it is a rationale or not. We use binary cross entropy between the predicted and the ground truth rationales to calculate the loss for token classification, denoted by $L_{rationale}$. This is different from the method illustrated in the following paper ~\cite{mathew2021hatexplain}, where they attempt to directly change the attention weights inside the model. Along with the rationale classifier, we add two more parallel classification layers-- one to classify labels (loss $L_{label}$), and another to classify targets (loss $L_{target}$). The classification of labels is a multi-class problem, whereas target classification is a multi-label problem. Furthermore, we also consider three classes (hate speech, offensive speech, and normal) and two classes (abusive or not abusive) as two variants for the label classification task. The final loss of the model ($L_{total}$) is shown in equation \ref{eq2}, where $\beta$ controls the impact of the rationales and $\gamma$ controls the impact of the targets. We denote this as the \textsc{BERT}-Rationale-Label-Target (\textsc{BERT-RLT}) classifier.
\begin{equation}
    \footnotesize
    L_{total}=L_{label}+\beta \cdot L_{rationale} + \gamma \cdot L_{target}
    \label{eq2}
\end{equation}

\subsection{Classification using rationales}

\begin{figure}[!htbp]
    \centering
    \includegraphics[width=0.7\columnwidth]{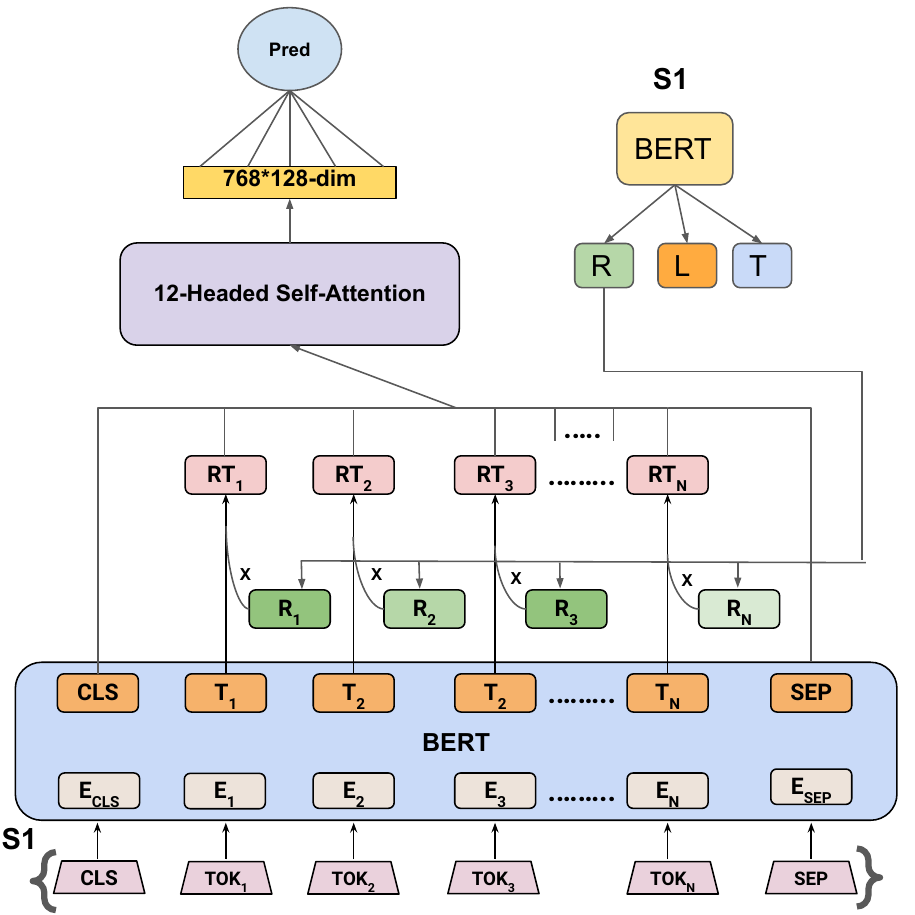}
    \caption{The RGFS-SA architecture. BERT-RLT is the model that predicts the rationales in a sentence $S1$, which are then added onto the BERT model using a self attention layer.}
    \label{fig:RGFS-Self Attention Architecture}
\end{figure}

\noindent\textsc{\texttt{RGFS-SA}}: Input sentences are first run through \textsc{BERT-RLT} (BERT-Rationale-Label-Target) to obtain probability scores for each token in the sentence being a positive rationale, i.e., a rationale belonging to the hate/abusive speech class. The inputs are then passed through \textsc{BERT}, and the last hidden state (LHS) output from \textsc{BERT} is updated by multiplying each token's output by its corresponding rationale score from the rationale-extractor, thereby assigning relative importance to each token with respect to other tokens for classification. A multi-headed self-attention layer is then constructed on top of the updated LHS; the updated LHS is taken as the query, key, and value vectors for attention. The resulting output is passed through a fully connected layer to generate final predictions. The weights of \textsc{BERT-RLT} (the rationale-extractor) are frozen and are not updated during the training of the classifier. We denote this classifier as \textsc{RGFS}-SelfAttention-Classifier (\textsc{RGFS-SA}). Figure~\ref{fig:RGFS-Self Attention Architecture} shows a schematic of this architecture.

\begin{figure}[!htbp]
    \centering
    \includegraphics[width=0.7\columnwidth]{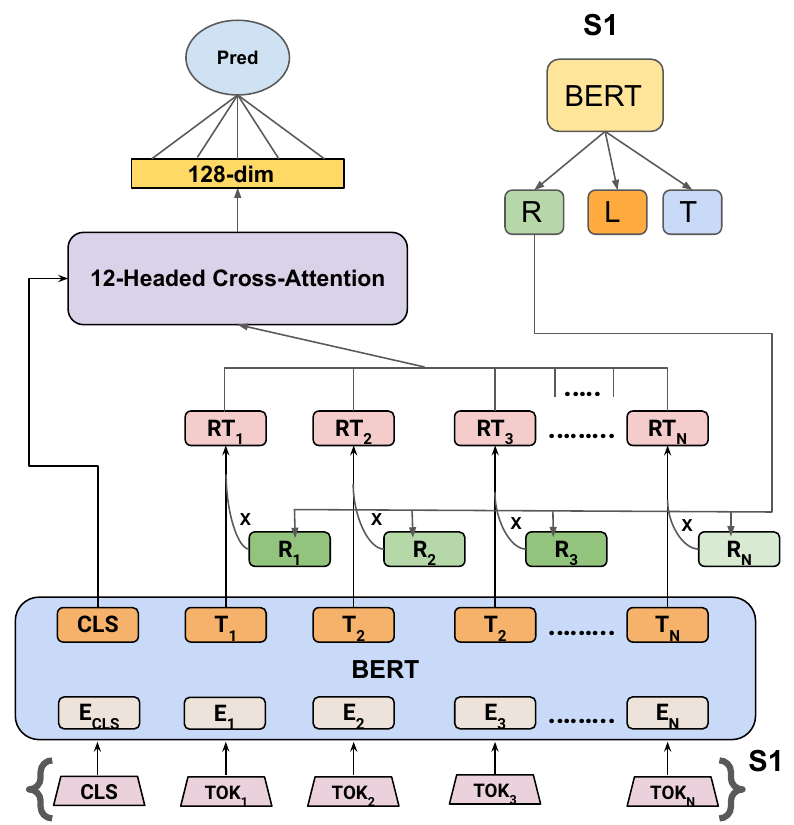}
    \caption{The RGFS-CA architecture, where the BERT-RLT model predicts the rationales for a sentence $S1$, which are then added onto the BERT model using a cross attention layer.}
    \label{fig:RGFS-Cross Attention Architecture}
\end{figure}

\noindent\textsc{\texttt{RGFS-CA}}: Instead of using self-attention on the updated LHS in the \textsc{RGFS-SA} model, we introduce a cross-attention layer between the CLS-pooled output obtained from \textsc{BERT} and the updated LHS here. We aim to have the model learn to get a representation of the complete sentence in its CLS-pooled output that works well with the architecture we add over \textsc{BERT} for the task. This CLS-pooled output is taken as the query vector, and the LHS*(token-wise rationale scores) is used as the key and value vectors. The weights of the rationale-extractor are kept frozen as in the previous case. This classifier is denoted as \textsc{RGFS}-CrossAttention-Classifier (\textsc{RGFS-CA}). Figure~\ref{fig:RGFS-Cross Attention Architecture} shows the schematic.

Note that the RGFS model was not trained on the past dataset or \textbf{HX}. RGFS only uses a rationale predictor (which was trained on \textbf{HX}) to predict the rationales on the cross-domain dataset. The classifier module of the RGFS model was trained with the target dataset mentioned in section~\ref{eval_datasets}.

\subsection{Metrics}
To evaluate the models, we rely on classification performance and explainability metrics. We use macro F1-score to measure classification performance, which is a standard metric for imbalanced datasets. 

For explainability, we use the framework given by DeYoung~\cite{deyoung2020eraser}, we evaluate plausibility and faithfulness of the models. \textit{Plausibility} refers to how convincing the model's interpretation is to humans, while \textit{faithfulness} aims to measure the reasoning of the model to arrive at a prediction~\cite{jacovi2020towards}.

\noindent\textbf{Plausibility}: To measure plausibility, we consider both discrete and soft selection metrics. As for discrete metrics, we report the \textit{IOU F1} and \textit{token F1} scores, and for soft selection we use \textit{AUCPRC} scores~\cite{deyoung2020eraser}. The token F1 is derived from token-wise precision and recall scores between predicted and ground truth rationales.
DeYoung~\cite{deyoung2020eraser} defines IOU at the token level -- for two spans, it is the size of the overlap of the tokens they cover divided by the size of their union. In IOU F1 metric, a prediction is considered a match if the overlap with any of the ground truth rationales is more than 0.5. These partial matches are used to calculate the IOU F1 score. Thus, while the token-level F1 score (token F1) measures the token level matching, IOU F1-score also awards credits to partial matches~\cite{everingham2010pascal}

\noindent\textbf{Faithfulness}: To measure faithfulness, we report two complementary metrics: \textit{comprehensiveness} and \textit{sufficiency}~\cite{deyoung2020eraser}.
\begin{compactitem}
    
    \item\textit{Comprehensiveness}: To measure comprehensiveness, we create a contrasting example $\tilde{x}_i$, for each post $x_i$,
    where $\tilde{x}_i$ is calculated by removing the predicted rationales $r_i$\footnote{We select the top 5 tokens as the rationales as per Mathew~\cite{mathew2021hatexplain}} from $x_i$. Let $m(x_i)_j$ be the original prediction probability provided by a model $m$ for the predicted class $j$. $m(x_i \backslash r_i)_j$ is then defined as the predicted probability of $\tilde{x}_i$ ($=x_i \backslash r_i$) by the model $m$ for the  class $j$.
    We would expect the model prediction to be lower on removing the rationales. We can measure this as: ${\textit{comprehensiveness}} = m(x_i)_j - m(x_i \backslash r_i)_j$.
    A high value of comprehensiveness implies that the rationales in the text are influential in the model's prediction.
    
    \item\textit{Sufficiency}: Measures the degree to which extracted rationales are adequate for a model to make a prediction. This can be measured as: ${\textit{sufficiency}} = m(x_i)_j - m(r_i)_j$. A low sufficiency score would imply that the model's performance on text containing just the rationales is close to the performance of the model on the complete text.
\end{compactitem}
\subsection{Rationale annotation}

In order to evaluate the predicted rationales on the target domain/dataset, we sample 50 datapoints from the abusive class (hateful/offensive) from the test splits of the five target datasets (FA, DA, OD, BA, and WH) and annotate the rationales in them.

Six annotators participated in the process, including 2 bachelors students and 4 PhD students. All the annotators were aged between 20-30 years, and they all had experience in the domain of abusive language research. We used Docanno\footnote{https://github.com/doccano/doccano}, an open source annotation platform to perform the annotation task. Each annotator was given a secure account and an interface where the posts were shown (see Appendix). Each post was labelled by two annotators. For each post, the annotators were required to classify the post as abusive or normal. If they found the post abusive, they had to mark rationales in the form of phrases in the text following the guidelines given by past research~\cite{mathew2021hatexplain}. We consider a post abusive when both annotators have marked it as abusive.

The average Jaccard overlap between the annotators across different datasets is 0.64. We also simulate random rationales and find the average Jaccard overlap to be around 0.30 (see in Appendix).

When tokenizing the post (having rationales), we divide the post into rationale and non-rationale phrases based on the rationale annotation. Then these phrases are individually tokenized. Now for all the tokens in rationale phrases, the rationale label is 1 while this is 0 for the tokens in non rationale phrases. In this way we generate a vector of 1's and 0's for each phrase, i.e., rationale vector. Finally, we concatenate the tokenized phrases and their corresponding rationale vectors.

\subsection{Experimental setup} 
The rationale extraction performance on the HateXplain dataset {as shown in Table~\ref{tab:hatexplain-performance}} is compared using the same \textit{train}:\textit{development}:\textit{test} splits of 8:1:1 as used by Mathew~\cite{mathew2021hatexplain}. For the evaluation datasets, we maintain a stratified split of \textit{train}:\textit{development}:\textit{test} in the ratio 7:1:2, similar to previous research on cross domain evaluation~\cite{swamy-etal-2019-studying}\footnote{The amount of data per split is noted in Appendix.}.  We use samples from the training dataset to train the few-shot detection models, the validation dataset is used to find the best model while training. Finally, we report the result on the test data.

We set the token length to 128 for reducing model size\footnote{We consider first 128 tokens when $>$ 128 tokens (0.05\% cases).}. All our results are reported on the test set of the corresponding dataset (either source or target). We also highlight the best performance using \textbf{bold} font in all the tables. In Tables \ref{tab:hatexplain-performance} and \ref{tab:few-shot}, we also show the second best using \underline{underline}. For cross domain few-shot evaluation, we use 50, 100, 150, and 200 training datapoints from each class to train the models in the new domains\footnote{None of the datasets have less than 200 datapoints for any label.}. We create five such different random sets of 50, 100, 150, and 200 datapoints for each target dataset to make our evaluations robust and we report average performance across these sets. 

For all the models trained, we vary the learning rate as the main hyperparameter, taking up the following values: $1e^{-5}$,  $3e^{-5}$ and $5e^{-5}$. With regards to equation~\ref{eq2}, we vary $\beta$ and $\gamma$ through $1,2,5,10,100$, achieving maximum performance on the validation set with $\beta=2$ and $\gamma=10$ for the \textsc{BERT-RLT} model. Once the best model is found for in-domain performance, we fix the model for all cross-domain evaluations. The learning rate for the cross domain few-shot classification models is fixed at $1e^{-5}$. 
Note that we did not explicitly provide a comparison with the earlier papers, since most of them use inferior models compared to BERT.

\section{Results}

\subsection{Rationale extraction}

We show the different variations to train the rationale classifier in Table~\ref{tab:hatexplain-performance}. At first, we consider the rationale classifier alone. This results in a macro F1 score of $0.71$. If we consider random rationales, we obtain a macro F1 score of $0.49$ which highlights the difficulty of predicting rationales. Next, we consider two variants of the label classification problem - (i) in the two class variant, we consider toxic and non-toxic as the final labels. (ii) In the three class variant, we consider - hate speech, offensive and normal - similar to the HateXplain paper.

We do not see an improvement in rationale classification performance when we add in the label classification task (model denoted by BERT-RL) along with rationale classification. We then add the target classification task as well, the new model being denoted by BERT-RLT. In the three class variant, the BERT-RLT model performs slightly better than the original rationale classification model. We get the best performance for BERT-RLT in the two class variant. We observe that \textsc{BERT-RLT} outperforms the base rationale classifier by 4 F1 points. Henceforth, we will use this BERT-RLT model as our rationale predictor. Across all the models, the rationale accuracy (R-Acc) is $>0.95$ since it is a highly imbalanced task. 

The performance on the label classification task remains almost the same for both the variants (two class and three class). For the three class variant, the marco F1 score (L-F1 in Table~\ref{tab:hatexplain-performance}) is in the range 0.67-0.69, whereas for the two class variant it is 0.77. The targets classification task is extremely difficult with 22 different classes (see in Appendix). For both two and three class variants of the BERT-RLT model, macro F1 score reaches around 0.21. We mostly resort to BERT models as we did not find much difference with other transformer models like RoBERTa (1\% difference in rationale classification F1 score).

\begin{table}
\centering
\caption{Performance of different models on the HateXplain test set. The models are denoted by BERT-X where the letters in X denote the tasks that particular variant uses. R denotes rationale, L denotes label, T denotes target classification. F1 denotes the macro F1-score and Acc denotes the accuracy. L-F1: label macro F1, L-Acc: label accuracy, R-F1: rationale macro F1 and R-Acc: rationale accuracy.}
\label{tab:hatexplain-performance}
\begin{tabular}{lcccc}
Model                & L-F1 & L-Acc & R-F1 & R-Acc \\\hline
\textsc{BERT-R-Random}  & --     & --     & 0.49      & 0.97   \\
\textsc{BERT-R}         & --     & --     & 0.66    & 0.97   \\\hline
\multicolumn{5}{c}{Two classes} \\\hline
\textsc{BERT-L}         & 0.77     & 0.78     & --    & --   \\
\textsc{BERT-RL}        & 0.77     & 0.78     & 0.66  & 0.96   \\
\textsc{BERT-RLT}       & 0.77     & 0.78     & \textbf{0.70} & \textbf{0.98}   \\\hline
\multicolumn{5}{c}{Three classes} \\\hline
\textsc{BERT-L}         & 0.69     & 0.70     & --    & --   \\
\textsc{BERT-RL}        & 0.68     & 0.69     & 0.64  & 0.96   \\
\textsc{BERT-RLT}       & 0.67     & 0.68     & \underline{0.68}  & \underline{0.97}   \\

\end{tabular}
\end{table}

\subsection{Similarity across different domains}
A well-known strategy to improve cross domain few-shot classification is to use a model trained on a similar domain. Following this strategy, we aim to select the best source among the datasets (excluding the HateXplain dataset). We first calculate the normalised term distribution for all the posts in a particular dataset. Next we use the pairwise cosine similarity between the term distribution of the two datasets~\cite{ruder2017learning}. For a particular target dataset, we first fine-tune the model using all the training points of the best source dataset (see the Table in Appendix for best source-target pairs). We name this model as BERT-L-DOM, where the DOM refers to the best source domain for that particular target dataset. We then add a new last layer on this model to train on the training samples of the target dataset. We also compare the similarity value of the target dataset with the HateXplain (HX) dataset and find there is an average drop of 25\% across different target datasets. Intuitively, this can be due to the different sources (Twitter and Gab) and a different timeline of data collection.

\begin{table}
\centering
\caption{Comparison of cross domain few-shot performance for \textsc{BERT-L}, \textsc{BERT-L-DOM}, \textsc{RGFS}-CA and \textsc{RGFS}-SA. Here, we train a particular model using few labeled datapoints and evaluate on the test data from the same domain. Each cell shows the average macro F1-score on the test dataset after running the model for five times with different sets of datapoints for fine-tuning. \textsc{BERT-L-DOM} is the best cross domain model based on the performance corresponding to each dataset.}
\begin{tabular}{lp{0.3cm}llllll}\hline
\multirow{2}{*}{\textbf{Model}}    & \multirow{2}{*}{\textbf{Data}}     & \multicolumn{5}{c}{\textbf{No. of training datapoints} (per label)} \\\cline{3-7}
                          &                           & 50     & 100   & 150  & 200  & All   \\\hline
\textsc{BERT-L}           & \multirow{4}{*}{DA}       & 0.590  & 0.609 & 0.671 & 0.674 & \textbf{0.768} \\
\textsc{BERT-L-DOM}           &                       &\textbf{0.656}  & \underline{0.682}  & 0.689 & 0.684 & 0.766 \\
\textsc{RGFS}-CA          &                           & 0.636  & 0.681 & \textbf{0.700} & \underline{0.699} & \underline{0.766}\\
\textsc{RGFS}-SA          &                           & \underline{0.653}  & \textbf{0.693} & \underline{0.695} & \textbf{0.711} & 0.752 \\\hline
\textsc{BERT-L}           & \multirow{4}{*}{OD}       & 0.588  & 0.651 & 0.689  & 0.703 & \textbf{0.785} \\
\textsc{BERT-L-DOM}           &                       & 0.561  & 0.637  & 0.693  & 0.703  & 0.776 \\
\textsc{RGFS}-CA          &                           & \underline{0.627}  & \underline{0.668} & \underline{0.702} & \underline{0.712} & 0.778 \\ 
\textsc{RGFS}-SA          &                           & \textbf{0.648}  & \textbf{0.686} & \textbf{0.710} & \textbf{0.725} & \underline{0.779} \\\hline
\textsc{BERT-L}           & \multirow{4}{*}{BA}       & 0.634  & 0.647 & 0.671 & 0.673 & \textbf{0.795} \\
\textsc{BERT-L-DOM}           &                       & \underline{0.650}  & 0.662  & \underline{0.684}  & \underline{0.700}  & 0.781 \\
\textsc{RGFS}-CA          &                           & 0.648  & \underline{0.673} & 0.683 & 0.690 & 0.787 \\ 
\textsc{RGFS}-SA          &                           & \textbf{0.658}  & \textbf{0.673} & \textbf{0.686} & \textbf{0.701} & \underline{0.791} \\\hline
\textsc{BERT-L}           & \multirow{4}{*}{FA}       & 0.597  & 0.653 & 0.677  & 0.687 & \underline{0.779} \\
\textsc{BERT-L-DOM}       &                           & \underline{0.655}  & 0.679  & 0.694  & \underline{0.702}  & 0.776 \\
\textsc{RGFS}-CA          &                           & 0.649  & \underline{0.691} & \textbf{0.707} & \textbf{0.709} & 0.773 \\ 
\textsc{RGFS}-SA          &                           & \textbf{0.665}  & \textbf{0.692} & \underline{0.700} & \textbf{0.709} & \textbf{0.784} \\\hline
\textsc{BERT-L}           & \multirow{4}{*}{WH}       & 0.692  & 0.733 & 0.757 & \underline{0.775} & \underline{0.857} \\
\textsc{BERT-L-DOM}       &                           & \textbf{0.755}  & \textbf{0.771}  & \textbf{0.780}  & \textbf{0.787}  & 0.844 \\
\textsc{RGFS}-CA          &                           & \underline{0.724}  & 0.735 & 0.752 & 0.767 & \textbf{0.858} \\ 
\textsc{RGFS}-SA          &                           & 0.719  & \underline{0.741} & \underline{0.757} & 0.772 & 0.846 \\\hline

\end{tabular}
\label{tab:few-shot}
\end{table}

\begin{table}
\caption{Examples where \textsc{RGFS} models perform better than {BERT-L-DOM}. The highlighted words are rationales identified within \textsc{RGFS} which enable it to classify correctly while \textsc{BERT-L-DOM} fails to do so. }
\label{tab:table-example1}
\begin{tabular}{p{0.6cm}p{0.6cm}p{1cm}p{4.5cm}}\hline
\scriptsize{\textsc{Ground Truth}} & \scriptsize{\textsc{RGFS-SA}} &\scriptsize{\textsc{BERT-L-DOM}} & \scriptsize{\textsc{Text}}\\\hline
hateful & hateful & normal & Wow they tried to give \hl{me} \hl{a} \hl{FEMALE} \hl{DOCTOR} TODAY. I \hl{didn't} even \hl{know} those EXISTEd… \\\hline
hateful & hateful & normal & user i am convinced the \hl{rapefugee} \hl{invasion} is an integral stage leading to this. considering their \hl{victims} tend to be \hl{minors}.\\\hline
\end{tabular}
\end{table}

\begin{table*}
\begin{center}
{\caption{Ablation study: Percentage drops in F1-scores upon using random rationales (RR) with RGFS models instead of rationales obtained from the BERT-RLT rationale predictor. Random rationale weights are assigned to non-rationales identified by BERT-RLT, while detected rationales are assigned low rationale weights.}\label{tab:ablation_results}
}
\begin{tabular}{ccc}
\hline
\textbf{Dataset} & \textbf{\% drop with RR Vs. RGFS-CA} & \textbf{\% drop with RR Vs. RGFS-SA} \\\hline
DA & -14.97\% & -13.49\% \\
OD & -13.10\% & -5.68\% \\
BA & -2.56\% & -4.18\% \\
FA & -3.87\% & -6.12\% \\
WH & -4.37\% & -3.08\% \\\hline
\end{tabular}
\end{center}

\end{table*}

\begin{table}
\centering
\caption{Average AUPRC, token-F1 and IOU-F1 scores for the rationales predicted by the models which were trained using different sets of 50 datapoints. For the models not utilising rationales in their architecture, LIME and SHAP are used to predict the rationales. \textsc{BERT-L-DOM} is the best cross-domain model for each dataset.}
\scriptsize
\begin{tabular}{lllll}\hline
\multirow{2}{*}{\textbf{Model}}    & \multirow{2}{*}{\textbf{Data}}     & \multicolumn{3}{c}{\textbf{Plausibility}} \\\cline{3-5}
                          &                           & AUPRC    & token-F1  & IOU-F1    \\\hline
\textsc{BERT-L-DOM + LIME}& \multirow{4}{*}{DA}       & 0.77  & 0.52 & 0.21 \\
\textsc{BERT-L-DOM + SHAP}&                           & 0.45  & 0.37 & 0.14 \\
\textsc{RGFS}-CA/SA       &                           &\textbf{0.84}  & \textbf{0.58} & \textbf{0.26} \\\hline
\textsc{BERT-L-DOM +LIME} & \multirow{4}{*}{OD}       & 0.49  & 0.36 & 0.10\\
\textsc{BERT-L-DOM +SHAP} &                           & 0.37  & 0.32 & 0.07\\
\textsc{RGFS}-CA/SA       &                           & \textbf{0.68}  & \textbf{0.54} & \textbf{0.11} \\\hline
\textsc{BERT-L-DOM + LIME}& \multirow{4}{*}{BA}       & 0.63  & 0.43 & 0.19 \\
\textsc{BERT-L-DOM +SHAP} &                           & 0.46  & 0.34 & 0.14 \\
\textsc{RGFS}-CA/SA       &                           & \textbf{0.76}  & \textbf{0.55} & \textbf{0.23} \\\hline
\textsc{BERT-L-DOM + LIME}& \multirow{4}{*}{FA}       & 0.61  & 0.46 & 0.11  \\
\textsc{BERT-L-DOM + SHAP}&                           & 0.48  & 0.36 & 0.06  \\
\textsc{RGFS}-CA/SA       &                           & \textbf{0.67}  & \textbf{0.54} & \textbf{0.13} \\\hline
\textsc{BERT-L-DOM + LIME}& \multirow{4}{*}{WH}       & 0.57  & 0.42 & 0.01 \\
\textsc{BERT-L-DOM + SHAP} &                           & 0.50  & 0.35 & 0.01 \\
\textsc{RGFS}-CA/SA       &                           & \textbf{0.84}  & \textbf{0.63} & 0.01  \\\hline

\end{tabular}
\label{tab:plausibility}
\end{table}

\subsection{Cross domain few-shot classification}

We perform cross domain few-shot classification by considering 5 random sets of $50$, $100$, $150$, $200$ samples per label from each training dataset and use it to train different model variants. As a baseline, we consider the \textsc{BERT}-L model which is finetuned directly on the target dataset. We compare this model with our proposed models - \textsc{RGFS}-SA and \textsc{RGFS}-CA which use an attention framework to include rationales in the architecture. 

When training with $50$ samples, both the RGFS models perform better than the \textsc{BERT}-L model (see Table~\ref{tab:few-shot}). In terms of macro F1 score, the \textsc{RGFS}-SA model outperforms the \textsc{BERT-L} by around 4 points. On the other hand, the \textsc{RGFS}-CA model outperforms the BERT-L by around 3.5 points. The highest difference between \textsc{RGFS}-SA/CA and BERT-L models appears for the FA dataset and the least for WH dataset. This suggests that the classification task for WH dataset might be easier compared to the FA dataset. The rationales predicted by the \textsc{BERT-RLT} help in the latter case to improve the classification in a cross domain few-shot setting. With increase in the number of datapoints, the difference between the \textsc{RGFS} models and the \textsc{BERT-L} models reduces but stays significant across different datasets except for the WH dataset. 

We also include the best cross domain model per dataset based on the similarity metric (see in Appendix). The best model trained on a source dataset (\textsc{BERT-L-DOM}) is trained again on this new dataset with a new last linear classification layer, on the number of the samples of the target dataset being used. The \textsc{RGFS-SA} model outperforms \textsc{BERT-L-DOM} in low data settings (50, 100 datapoints) for the OD, BA and FA datasets. In fact, with 100 training points, the \textsc{RGFS-SA} model beats the \textsc{BERT-L-DOM} model for all datasets except WH (some examples where \textsc{BERT-L-DOM} fails are noted in Table~\ref{tab:table-example1}). In addition, we also apply the few shot detection method used in~\cite{aluru2021deep} where we consider all the other datasets source dataset except the target dataset. We find that on average the BERT-L-DOM approach is 2\% ($\sigma=0.11$) better than this method. Hence we use the \textsc{BERT-L-DOM} for further experiments. Finally, the \textsc{RGFS-SA} models outperform these models by 4\% ($\sigma=0.06$) on average.

\noindent\textbf{Are the predicted rationales useful?} Here, we attempt to understand the impact of the predicted rationales in the overall RGFS architecture. Our approach is to dilute the importance of the rationales and observe how this affects the overall perfromance of the model. The rationale predictor predicts a score per token (can be $+ve/-ve$), where a higher score means more propensity to become a rationale. We select a threshold for each dataset based on the higher quartile value in the score distribution. We assume tokens with scores higher than the threshold (i.e., the top 25 percentile scores) represent the rationales and update their scores to a very low value $(-4)$\footnote{The value of 4 is based on the observed distribution.} to reduce their importance.

The rest of the tokens are assigned a random score based on a uniform distribution -- $\mathcal{U}_{(-5,5)}$\footnote{The value of 5 is based on the observed distribution.}. This modified vector is passed through softmax function to create the random rationale vector and the experiments with 50 datapoints in Table~\ref{tab:few-shot} is repeated. We observe that for this change there is an average percentage drop of around 6\% in F1 score for RGFS-CA and RGFS-SA respectively, with the highest drop being observed for the Davidson dataset. We also add a qualitative analysis of how the rationales help in the Appendix.

\begin{table}[!t]
\centering
\caption{Average comprehensiveness scores (Comp) and the sufficiency scores (Suff) for the rationales predicted by the models which were trained using different sets of 50 datapoints. For the models not utilising rationales in their architecture, LIME and SHAP are used to predict the rationales. \textsc{BERT-L-DOM} is the best cross domain model corresponding to each dataset. For sufficiency scores, lower values are better.}
\begin{tabular}{llcc}\hline
\multirow{2}{*}{\textbf{Model}}   & \multirow{2}{*}{\textbf{Data}}     & \multicolumn{2}{l}{\textbf{Faithfulness}} \\\cline{3-4}
                                &                     & Suff.($\downarrow$)    & Comp.\\\hline
\textsc{BERT-L-DOM + LIME}       & \multirow{5}{*}{DA}& -0.03  & \textbf{0.67}  \\
\textsc{BERT-L-DOM + SHAP}       &                    & \textbf{-0.29}  & -0.14  \\
\textsc{RGFS}-CA          &                           & -0.06  & 0.11  \\
\textsc{RGFS}-SA          &                           & -0.08  & 0.25 \\\hline
\textsc{BERT-L-DOM + LIME}       & \multirow{5}{*}{OD}& -0.02  & 0.09 \\
\textsc{BERT-L-DOM + SHAP}       &                    & -0.10  & -0.09 \\
\textsc{RGFS}-CA          &                           & -0.04  & 0.04  \\
\textsc{RGFS}-SA          &                           & -0.08  & \textbf{0.29} \\\hline
 \\
\textsc{BERT-L-DOM + LIME}       & \multirow{5}{*}{BA}& -0.05  & \textbf{0.34} \\
\textsc{BERT-L-DOM + SHAP}       &                    & -0.38  & -0.38 \\
\textsc{RGFS}-CA          &                           & -0.04  & 0.06  \\
\textsc{RGFS}-SA          &                           & \textbf{-0.07}  & 0.19 \\\hline
\textsc{BERT-L-DOM + LIME}       & \multirow{5}{*}{FA}& -0.02  & \textbf{0.40} \\
\textsc{BERT-L-DOM + SHAP}       &                    & -0.09  & 0.09 \\
\textsc{RGFS}-CA          &                           & \textbf{-0.13}  & 0.09  \\
\textsc{RGFS}-SA          &                           & -0.11  & 0.26 \\\hline
\textsc{BERT-L-DOM + LIME} & \multirow{5}{*}{WH}       & -0.07  & \textbf{0.20} \\
\textsc{BERT-L-DOM + SHAP}       &                    & -0.09  & -0.09 \\
\textsc{RGFS}-CA          &                           & 0.01   & 0.03  \\
\textsc{RGFS}-SA          &                           & 0.06   & 0.06 \\\hline

\end{tabular}

\label{tab:faithfulness}
\end{table}

\subsection{Explainability}
\label{Explainability}

In order to evaluate the predicted rationales, we take the models which are domain adapted using 5 different random sets of 50 datapoints (per dataset) from the training dataset\footnote{Corresponding to column $1$ in Table \ref{tab:few-shot}.}, and the rationale annotated data (see in Appendix). To compare with models which do not predict rationales within them, we pass the model output through two explainability methods namely,  LIME~\cite{lime} and SHAP~\cite{NIPS2017_7062} to get importance scores for each word. After getting the raw vector of importance scores, we  normalise the score between 0 and 1 using min-max normalisation. For the \textsc{RGFS} models, we perform the same operation but with the rationales predicted by the \textsc{BERT-RLT}. Next, we evaluate these rationales.

\noindent\textbf{Plausibility}: For plausibility, we first consider the soft token metric - AUPRC. The rationales used in RGFS models outperform the \textsc{BERT-L-DOM + LIME} configuration by 2 points on average and \textsc{BERT-L-DOM + SHAP} configuration by 4 points on average (see Table~\ref{tab:plausibility}). The difference in some of the datasets (BA, WH, OD) is more significant than the others (DA and FA dataset). In terms of token F1, \textsc{RGFS} outperforms the \textsc{BERT-L-DOM + LIME} configuration by around 1.5 F1 points and \textsc{BERT-L-DOM + SHAP} around 2 F1 points (see Table~\ref{tab:plausibility}). The results are closer for the IOU scores since it also considers partial matches but here again the predicted rationales outperform LIME/SHAP rationales across all datasets except for WH. Overall, we observe that the rationales predicted using the BERT-RLT model provide more plausible explanations that the rationales generated using LIME/SHAP across all the target datasets. 

\noindent\textbf{Faithfulness}: We measure the faithfulness of the rationales using \textit{sufficiency} and \textit{comprehensiveness} to understand whether the rationales act as correct reasoning behind the model predictions. In terms of sufficiency, we observe that the \textsc{RGFS-SA} model outperforms the \textsc{RGFS-CA} across DA, OD, BA datasets (see Table~\ref{tab:faithfulness}). The \textsc{RGFS} based models overall outperform the \textsc{BERT-L-DOM + LIME} configuration across all datasets except WH but underperforms when compared to the SHAP based configuration. In terms of comprehensiveness, \textsc{BERT-L-DOM} is slightly superior to the \textsc{RGFS-SA} model. Additional artefacts in the model might be a reason for this slight underperformance. SHAP based rationales are much worse compared to other variations in terms of comprehensiveness.

\section{Conclusion}

In the field of abusive language, one of the major issues in building better detection is the variability in the context across demographies and the subjective nature of abuse. Further, since platforms are coming up with stronger moderation algorithms, the language of abuse is also changing to bypass such moderation. The paper~\cite{10.1145/3415163} discussed in detail the temporal nature of hate speech. Hence, many researchers are aiming to create dynamic datasets~\cite{vidgen2020learning}, since static datasets cannot serve as consistent benchmarks for abusive language research. In addition to such dynamic datasets, we need models that can learn using a few samples from target domains and are more explainable in order to identify the drawbacks of the models in a new domain.

In this paper, we show that models that utilise rationales can perform better in cross domain few-shot classification than the models without them. These models also provide competitive performance as compared to the best model pre-trained on a source dataset (BERT-L-DOM) , which has already been trained using more abusive examples. Further, the rationale predictors of models like \textsc{RGFS} provide more plausible explanations compared to traditional LIME/SHAP-based explanations, whilst slightly under-performing in terms of faithfulness. In future, we would focus on creating a typology of the errors in classification and rationale extraction task. We also plan to extend this framework to other languages.

\bibliography{ecai}

\begin{thebibliography}{10}

\bibitem{aluru2021deep}
Sai~Saketh Aluru, Binny Mathew, Punyajoy Saha, and Animesh Mukherjee, `A deep
  dive into multilingual hate speech classification', in {\em ECML PKDD 2020},
  pp. 423--439. Springer International Publishing, (2021).

\bibitem{awal2020analyzing}
Md~Rabiul Awal, Rui Cao, Roy Ka-Wei Lee, and Sandra Mitrovi{\'c}, `On analyzing
  annotation consistency in online abusive behavior datasets', {\em arXiv
  preprint arXiv:2006.13507}, (2020).

\bibitem{basile-etal-2019-semeval}
Valerio Basile, Cristina Bosco, Elisabetta Fersini, Debora Nozza, Viviana
  Patti, Francisco~Manuel Rangel~Pardo, Paolo Rosso, and Manuela Sanguinetti,
  `{S}em{E}val-2019 task 5: Multilingual detection of hate speech against
  immigrants and women in {T}witter', in {\em SemEval 2019}, pp. 54--63.
  Association for Computational Linguistics, (June 2019).

\bibitem{baziotis-pelekis-doulkeridis:2017:SemEval2}
Christos Baziotis, Nikos Pelekis, and Christos Doulkeridis, `Datastories at
  semeval-2017 task 4: Deep lstm with attention for message-level and
  topic-based sentiment analysis', in {\em SemEval-2017}, pp. 747--754,
  Vancouver, Canada, (August 2017). Association for Computational Linguistics.

\bibitem{bhat2021self}
Meghana~Moorthy Bhat, Alessandro Sordoni, and Subhabrata Mukherjee,
  `Self-training with few-shot rationalization: Teacher explanations aid
  student in few-shot nlu', {\em arXiv preprint arXiv:2109.08259}, (2021).

\bibitem{ijcai2022p398}
Ziqiu Chi, Zhe Wang, Mengping Yang, Wei Guo, and Xinlei Xu, `Better embedding
  and more shots for few-shot learning', in {\em Proceedings of the
  Thirty-First International Joint Conference on Artificial Intelligence,
  {IJCAI-22}}, ed., Lud~De Raedt, pp. 2874--2880. International Joint
  Conferences on Artificial Intelligence Organization, (7 2022).
\newblock Main Track.

\bibitem{chu2020survey}
Chenhui Chu and Rui Wang, `A survey of domain adaptation for machine
  translation', {\em Journal of Information Processing}, {\bf 28},  413--426,
  (2020).

\bibitem{confalonieri2019trepan}
Roberto Confalonieri, Tillman Weyde, Tarek~R Besold, and Ferm{\'\i}n Moscoso
  del~Prado Mart{\'\i}n, `Trepan reloaded: a knowledge-driven approach to
  explaining artificial neural networks', {\em arXiv preprint
  arXiv:1906.08362}, (2019).

\bibitem{danilevsky2020survey}
Marina Danilevsky, Kun Qian, Ranit Aharonov, Yannis Katsis, Ban Kawas, and
  Prithviraj Sen, `A survey of the state of explainable ai for natural language
  processing', {\em arXiv preprint arXiv:2010.00711}, (2020).

\bibitem{davidson-etal-2019-racial}
Thomas Davidson, Debasmita Bhattacharya, and Ingmar Weber, `Racial bias in hate
  speech and abusive language detection datasets', in {\em Proceedings of the
  Third Workshop on Abusive Language Online}, pp. 25--35, Florence, Italy,
  (August 2019). Association for Computational Linguistics.

\bibitem{davidson}
Thomas Davidson, Dana Warmsley, Michael Macy, and Ingmar Weber, `Automated hate
  speech detection and the problem of offensive language', ICWSM '17, pp.
  512--515, (2017).

\bibitem{devlin-etal-2019-bert}
Jacob Devlin, Ming-Wei Chang, Kenton Lee, and Kristina Toutanova, `{BERT}:
  Pre-training of deep bidirectional transformers for language understanding',
  in {\em NAACL}, pp. 4171--4186, Minneapolis, Minnesota, (June 2019). ACL.

\bibitem{deyoung2020eraser}
Jay DeYoung, Sarthak Jain, Nazneen~Fatema Rajani, Eric Lehman, Caiming Xiong,
  Richard Socher, and Byron~C Wallace, `Eraser: A benchmark to evaluate
  rationalized nlp models', in {\em Proceedings of the 58th Annual Meeting of
  the Association for Computational Linguistics}, pp. 4443--4458, (2020).

\bibitem{elsherief2018hate}
Mai ElSherief, Vivek Kulkarni, Dana Nguyen, William~Yang Wang, and Elizabeth
  Belding, `Hate lingo: A target-based linguistic analysis of hate speech in
  social media', in {\em Proceedings of the International AAAI Conference on
  Web and Social Media}, volume~12, (2018).

\bibitem{everingham2010pascal}
Mark Everingham, Luc Van~Gool, Christopher~KI Williams, John Winn, and Andrew
  Zisserman, `The pascal visual object classes (voc) challenge', {\em
  International journal of computer vision}, {\bf 88}(2),  303--338, (2010).

\bibitem{misogyny}
Elisabetta Fersini, Debora Nozza, and Paolo Rosso, `Overview of the evalita
  2018 task on automatic misogyny identification (ami)', {\em EVALITA
  Evaluation of NLP and Speech Tools for Italian}, {\bf 12}, ~59, (2018).

\bibitem{fortuna-etal-2020-toxic}
Paula Fortuna, Juan Soler, and Leo Wanner, `Toxic, hateful, offensive or
  abusive? what are we really classifying? an empirical analysis of hate speech
  datasets', in {\em LREC}, pp. 6786--6794, Marseille, France, (May 2020).
  European Language Resources Association.

\bibitem{FORTUNA2021102524}
Paula Fortuna, Juan Soler-Company, and Leo Wanner, `How well do hate speech,
  toxicity, abusive and offensive language classification models generalize
  across datasets?', {\em Information Processing and Management}, {\bf 58}(3),
  102524, (2021).

\bibitem{founta2018large}
Antigoni-Maria Founta, Constantinos Djouvas, Despoina Chatzakou, Ilias
  Leontiadis, Jeremy Blackburn, Gianluca Stringhini, Athena Vakali, Michael
  Sirivianos, and Nicolas Kourtellis, `Large scale crowdsourcing and
  characterization of twitter abusive behavior', in {\em ICWSM 2018}. AAAI
  Press, (2018).

\bibitem{gillespie2018custodians}
Tarleton Gillespie, {\em Custodians of the Internet: Platforms, content
  moderation, and the hidden decisions that shape social media}, Yale
  University Press, 2018.

\bibitem{dontstoppretraining2020}
Suchin Gururangan, Ana Marasović, Swabha Swayamdipta, Kyle Lo, Iz~Beltagy,
  Doug Downey, and Noah~A. Smith, `Don't stop pretraining: Adapt language
  models to domains and tasks', in {\em Proceedings of ACL}, (2020).

\bibitem{jacovi2020towards}
Alon Jacovi and Yoav Goldberg, `Towards faithfully interpretable {NLP} systems:
  How should we define and evaluate faithfulness?', in {\em Proceedings of the
  58th Annual Meeting of the Association for Computational Linguistics}, pp.
  4198--4205, Online, (2020). Association for Computational Linguistics.

\bibitem{Dehumani67:online}
Aliza Luft.
\newblock Dehumanization and the normalization of violence: It’s not what you
  think – items.
\newblock \url{https://tinyurl.com/s4mxz7j2}, May 2019.
\newblock (Accessed on 05/02/2021).

\bibitem{NIPS2017_7062}
Scott~M Lundberg and Su-In Lee, `A unified approach to interpreting model
  predictions', in {\em Advances in Neural Information Processing Systems 30},
  eds., I.~Guyon, U.~V. Luxburg, S.~Bengio, H.~Wallach, R.~Fergus,
  S.~Vishwanathan, and R.~Garnett,  4765--4774, Curran Associates, Inc.,
  (2017).

\bibitem{10.1145/3415163}
Binny Mathew, Anurag Illendula, Punyajoy Saha, Soumya Sarkar, Pawan Goyal, and
  Animesh Mukherjee, `Hate begets hate: A temporal study of hate speech', {\em
  Proc. ACM Hum.-Comput. Interact.}, {\bf 4}(CSCW2), (oct 2020).

\bibitem{mathew2021hatexplain}
Binny Mathew, Punyajoy Saha, Seid~Muhie Yimam, Chris Biemann, Pawan Goyal, and
  Animesh Mukherjee, `Hatexplain: A benchmark dataset for explainable hate
  speech detection', in {\em Proceedings of the AAAI Conference on Artificial
  Intelligence}, volume~35, pp. 14867--14875, (2021).

\bibitem{melamud-etal-2019-combining}
Oren Melamud, Mihaela Bornea, and Ken Barker, `Combining unsupervised
  pre-training and annotator rationales to improve low-shot text
  classification', in {\em EMNLP-IJCNLP}, pp. 3884--3893, Hong Kong, China,
  (November 2019). ACL.

\bibitem{mishra2019tackling}
Pushkar Mishra, Helen Yannakoudakis, and Ekaterina Shutova, `Tackling online
  abuse: A survey of automated abuse detection methods', {\em arXiv preprint
  arXiv:1908.06024}, (2019).

\bibitem{mishra2021user}
Pushkar Mishra, Helen Yannakoudakis, and Ekaterina Shutova, `The user behind
  the abuse: A position on ethics and explainability', {\em arXiv preprint
  arXiv:2103.17191}, (2021).

\bibitem{ijcai2022p102}
Runliang Niu, Zhepei Wei, Yan Wang, and Qi~Wang, `Attexplainer: Explain
  transformer via attention by reinforcement learning', in {\em Proceedings of
  the Thirty-First International Joint Conference on Artificial Intelligence,
  {IJCAI-22}}, ed., Lud~De Raedt, pp. 724--731. International Joint Conferences
  on Artificial Intelligence Organization, (7 2022).
\newblock Main Track.

\bibitem{lime}
Marco~Tulio Ribeiro, Sameer Singh, and Carlos Guestrin, `"why should {I} trust
  you?": Explaining the predictions of any classifier', in {\em ACM SIGKDD},
  pp. 1135--1144, (2016).

\bibitem{Ro_Rist_Carbonell_Cabrera_Kurowsky_Wojatzki_2016}
Björn Roß, Michael Rist, Guillermo Carbonell, Benjamin Cabrera, Nils
  Kurowsky, and Michael~Maximilian Wojatzki.
\newblock Measuring the reliability of hate speech annotations: The case of the
  european refugee crisis, Nov 2016.

\bibitem{ruder2017learning}
Sebastian Ruder and Barbara Plank, `Learning to select data for transfer
  learning with bayesian optimization', {\em arXiv preprint arXiv:1707.05246},
  (2017).

\bibitem{saito2019semi}
Kuniaki Saito, Donghyun Kim, Stan Sclaroff, Trevor Darrell, and Kate Saenko,
  `Semi-supervised domain adaptation via minimax entropy', {\em ICCV}, (2019).

\bibitem{schmidt2021explainability}
Robin~M Schmidt, `Explainability-aided domain generalization for image
  classification', {\em arXiv preprint arXiv:2104.01742}, (2021).

\bibitem{stappen2020cross}
Lukas Stappen, Fabian Brunn, and Bj{\"o}rn Schuller, `Cross-lingual zero-and
  few-shot hate speech detection utilising frozen transformer language models
  and axel', {\em arXiv preprint arXiv:2004.13850}, (2020).

\bibitem{swamy-etal-2019-studying}
Steve~Durairaj Swamy, Anupam Jamatia, and Bj{\"o}rn Gamb{\"a}ck, `Studying
  generalisability across abusive language detection datasets', in {\em CoNLL},
  pp. 940--950, Hong Kong, China, (November 2019). Association for
  Computational Linguistics.

\bibitem{tenney-etal-2020-language}
Ian Tenney, James Wexler, Jasmijn Bastings, Tolga Bolukbasi, Andy Coenen,
  Sebastian Gehrmann, Ellen Jiang, Mahima Pushkarna, Carey Radebaugh, Emily
  Reif, and Ann Yuan, `The language interpretability tool: Extensible,
  interactive visualizations and analysis for {NLP} models', in {\em EMNLP},
  pp. 107--118, Online, (October 2020). Association for Computational
  Linguistics.

\bibitem{vedeler2019hate}
Janikke~Solstad Vedeler, Terje Olsen, and John Eriksen, `Hate speech harms: A
  social justice discussion of disabled norwegians? experiences', {\em
  Disability \& Society}, {\bf 34}(3),  368--383, (2019).

\bibitem{vidgen2020directions}
Bertie Vidgen and Leon Derczynski, `Directions in abusive language training
  data: Garbage in, garbage out', {\em arXiv preprint arXiv:2004.01670},
  (2020).

\bibitem{vidgen2020learning}
Bertie Vidgen, Tristan Thrush, Zeerak Waseem, and Douwe Kiela, `Learning from
  the worst: Dynamically generated datasets to improve online hate detection',
  {\em arXiv preprint arXiv:2012.15761}, (2020).

\bibitem{TheState71:online}
Emily~A Vogels.
\newblock The state of online harassment | pew research center.
\newblock
  \url{https://www.pewresearch.org/internet/2021/01/13/the-state-of-online-harassment/},
  2021.
\newblock (Accessed on 05/02/2021).

\bibitem{waseem-hovy-2016-hateful}
Zeerak Waseem and Dirk Hovy, `Hateful symbols or hateful people? predictive
  features for hate speech detection on {T}witter', in {\em Proceedings of the
  {NAACL} Student Research Workshop}, pp. 88--93, San Diego, California, (June
  2016). Association for Computational Linguistics.

\bibitem{Waseem2018}
Zeerak Waseem, James Thorne, and Joachim Bingel, {\em Bridging the Gaps: Multi
  Task Learning for Domain Transfer of Hate Speech Detection},  29--55,
  Springer International Publishing, Cham, 2018.

\bibitem{zaidan-etal-2007-using}
Omar Zaidan, Jason Eisner, and Christine Piatko, `Using {``}annotator
  rationales{''} to improve machine learning for text categorization', in {\em
  Human Language Technologies 2007: NAACL}, pp. 260--267, Rochester, New York,
  (April 2007). ACL.

\bibitem{zampieri-etal-2019-predicting}
Marcos Zampieri, Shervin Malmasi, Preslav Nakov, Sara Rosenthal, Noura Farra,
  and Ritesh Kumar, `Predicting the type and target of offensive posts in
  social media', in {\em NAACL}, pp. 1415--1420, Minneapolis, Minnesota, (June
  2019). ACL.

\bibitem{10.1145/3437963.3441758}
Zijian Zhang, Koustav Rudra, and Avishek Anand, `Explain and predict, and then
  predict again', in {\em WSDM}, WSDM '21, p. 418–426, New York, NY, USA,
  (2021). Association for Computing Machinery.

\end{thebibliography}

\appendix

\section{More details about the datasets}
\subsection{The HateXplain dataset (HX)}
The HateXplain dataset introduced by Mathew~\cite{mathew2021hatexplain} is a large dataset of 20k posts from Twitter and Gab. At the top-level, each post is annotated by three annotators into one of three categories -- \textit{hate speech}, \textit{offensive}, or \textit{normal}. Furthermore, groups or communities targeted in the post are also marked.
For abusive content (i.e. hate speech or offensive), tokens or word spans explaining abusive content are marked as rationales.  In total, there are 253 unique annotators as reported by Mathew~\cite{mathew2021hatexplain}. We use this dataset to train the rationale extraction model. The following is how we aggregated the ground truth for each type of annotation for this dataset:

\begin{compactitem}
    \item\textbf{Labels}: The final label for each datapoint is selected based on the majority label from the labels provided by three different annotators. We also convert this to a two-class problem by considering both hate speech and offensive labels in the `abusive' class and the normal label in the `non-abusive' class. We then follow the same majority selection criteria to select the final label. 
    \item\textbf{Rationales}: To provide the model with rationales as feedback, we convert the rationales by each annotator into Boolean vectors. Values in these Boolean vectors are 1 when the corresponding token (word) in the text is a part of a rationale. To create the \textbf{ground truth rationales}, we consider each token in the text and call it a rationale if at least two annotators have highlighted it as a rationale. The final ground truth rationales are Boolean vectors, considering the above constraint.
    \item\textbf{Targets}: For ground truth targets, we consider those \textbf{targets} that are labeled so by at least 2 annotators, after which we ignore those targets that appear less than 20 times in the complete dataset and replace them all with - `Others'. We find 22 targets which are noted in Table~\ref{tab:targets}.
\end{compactitem}

\if{0}\subsection{What are rationales?}

 In order to analyze the rationales that are important for abuse detection, we consider the phrases, i.e., spans of consecutive words which are marked by the annotators as the reasons for calling a post abusive in the HateXplain (HX) dataset. We then calculate the frequency distribution of these rationales. The top 5 frequent rationales include common slur/derogatory words like ‘ni**er, bi**h, k**e, ass. We would like to note that around 93\% of the 6021 unique rationale phrases have a count of 1. Examples of some rationales include “can not speak properly lack basic knowledge of biology”, “jew is just a ni**er turned inside out” and “evil mu*rat cult” etc. Average length of rationale phrases is around 6 words, hence many of the rationales are multi-word.\fi

 

\subsection{The Founta et al. dataset (FA)}
Founta~\cite{founta2018large} made available a large-scale Twitter dataset containing 4 different labels: \textit{hateful}, \textit{abusive}, \textit{normal} and \textit{spam}. Their work focused on dealing with the class imbalance in random samples from Twitter by filtering tweets in an incremental and iterative process, aided with boosted sampling. The quality of judgment was ensured by measuring agreement for over 20 annotators per tweet. The dataset contains 100k tweets and is the largest dataset considered in this paper. We ignore the datapoints annotated as spam from our analysis.

\subsection{The Davidson et al. dataset (DA)}
This work on automatic hate speech detection by Davidson~\cite{davidson} released a dataset of 24k tweets. Each tweet was queried from Twitter using a lexicon derived from Hatebase.org\footnote{www.hatebase.org}. Annotation was carried out by majority vote of at least three CrowdFlower workers. There are three labels in this dataset: \textit{hate speech}, \textit{offensive} and \textit{normal}. The high prevalence of abusive tweets is attributed in part to racial bias by Davidson~\cite{davidson-etal-2019-racial} who demonstrated that a classifier trained on the dataset shows significantly higher tendency to mark tweets written by African-Americans as abusive.

\subsection{The OLID dataset (OD)}
The Offensive Language Identification Dataset (OLID) dataset released by zampieri~\cite{zampieri-etal-2019-predicting} in the SemEval-2019 Task 6 (OffensEval) uses a modern hierarchical labelling scheme, where at the top-level, a tweet is classified as either offensive or not offensive. Tweets which are offensive are further divided into sub-categories based on whether the offense is untargeted or targeted against a group or individual. Similar to Davidson~\cite{davidson}, they employ a majority voting scheme to annotate tweets using the crowd-sourcing platform Appen\footnote{https://appen.com/}. For our work, we chose the 14k tweets from their top-level of annotation.

\subsection{The Basile et al. dataset (BA)} 
This hate speech dataset was used in the SemEval-2019 Task 5: Multilingual Detection of Hate Speech Against Immigrants and Women in Twitter \cite{basile-etal-2019-semeval}. To build this dataset, the authors monitored victims of known abusive accounts on Twitter and used important keywords and hashtags to filter their tweets. Some of the frequent keywords collected in the 13k English tweets are: \textit{migrant, refugee, \#buildthatwall, b*tch, women}. The tweets targeted against women were collected from a previous challenge on misogyny identification \cite{misogyny}.


\subsection{The Waseem and Hovy dataset (WH) }
Waseem~\cite{waseem-hovy-2016-hateful} published a hate speech detection dataset of 16k tweets. Their corpus is built by searching for slurs targeted against religious, sexual, gender, and ethnic minorities on Twitter. The authors manually annotated the tweets into one of three classes: \textit{racism}, \textit{sexism} or \textit{normal}. The 1,972 tweets in the \textit{racism} class are from just 6 users, and the 3,383 tweets in the \textit{sexism} class are from 613 users. While the original dataset was composed of 36\% positive (racism or sexism) labels, many of these tweets have since been taken down from Twitter. We managed to collect 10,018 tweets and ignored the racism class for further analysis as it only contained 13 datapoints.



\section{Interface for annotation}
\begin{figure}[!htpb]
    \centering
    \includegraphics[width=0.6\linewidth]{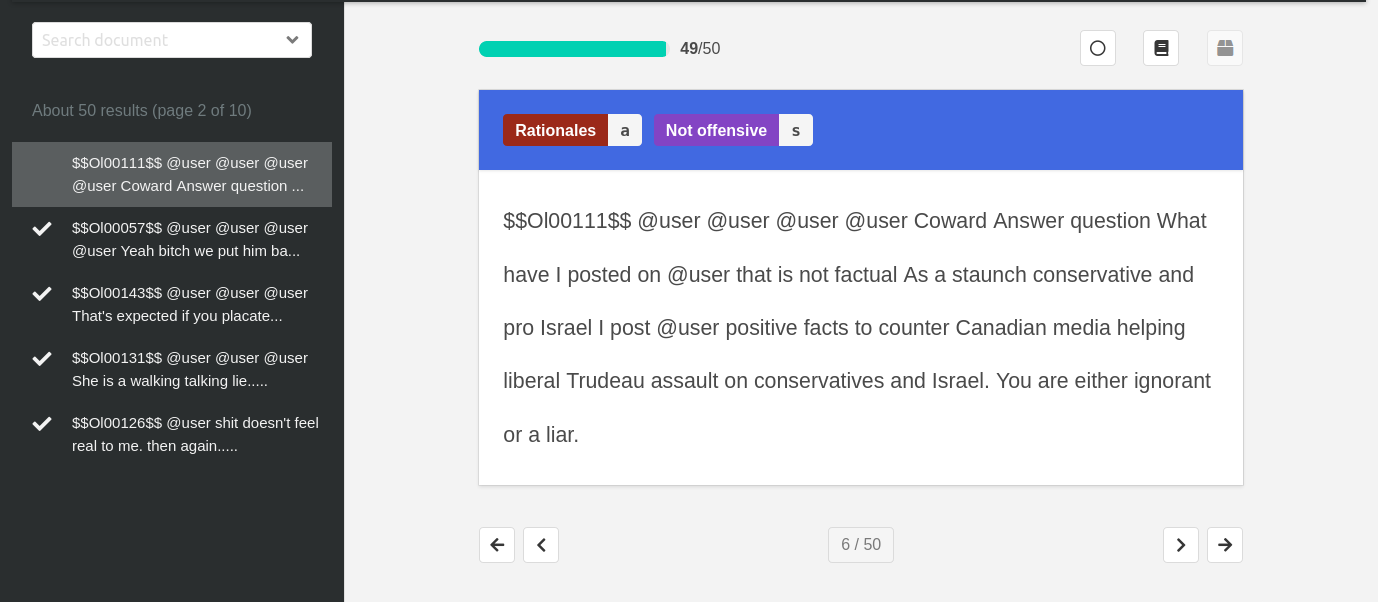}
    \caption{Interface for annotation.}
    \label{fig:interface}
\end{figure}

The interface for annotation that appear for each of the annotators is shown in Figure~\ref{fig:interface}.

\section{Other hyperparameters}
We set the batch size as 16 and train all models for 20 epochs, reporting the test performance at the epoch where the validation performance is the best. We used AdamW optimizer for optimization with default parameters. These values are constant through the whole experiment. For all the models, we used a dropout of 0.2 in the final linear layer, while for the RAFT models we further use 0.2 as dropout in the attention layer as well. Other than that for LIME~\cite{lime}, we used 10 features for explanation and 100 as the size of neighbourhood to learn the linear model. Other parameters were set to default. We evaluate the LIME/SHAP based explanation using the explanations for the most confident abusive class.
\section{Most similar source-target pairs}
All the pairwise cosine similarities are noted in Table ~\ref{tab:cross-domain-sim}. For each row in this table, we select the best source dataset based on the maximum similarity value.

\begin{table}[]
\scriptsize
\centering
\begin{tabular}{llllll|l}\hline
\begin{tabular}[c]{@{}l@{}}
Source $\rightarrow$ \\ Target $\downarrow$ \end{tabular} & DA & FA & OD & BA & WH & HX \\\hline
DA & -- & \textbf{0.76} & 0.68 & 0.73 & 0.73 & 0.56\\
FA   & 0.76 & -- & 0.80 & \textbf{0.92} & 0.87 & 0.68\\
OD     & 0.68 & 0.80 & -- & 0.82 & \textbf{0.84} & 0.62 \\
BA   & 0.73 & \textbf{0.92} & 0.82 & -- & 0.87 & 0.77\\
WH     & 0.73 & 0.87 & 0.84 & \textbf{0.88} & -- & 0.63\\
\end{tabular}
\caption{\scriptsize{In this table, we show the pairwise cosine similarity in the term distribution of different corpus. The row headers represent the target domain and the column header denote the source domain. For each row we select the best domain to transfer from.}}
\label{tab:cross-domain-sim}
\end{table}

\section{System description}
For all the experiments in this paper, we used a 48-core Xeon processor Linux based system with 126 GB RAM. For training  the  neural networks, we used 2 NVIDIA P100 GPUs having 16 GB RAM each with CUDA version 10.1. We primarily based our system on Python libraries. For preprocessing we used the ekphrasis\footnote{https://github.com/cbaziotis/ekphrasis} library. Huggingface's transformers library was used for BERT-based models, with PyTorch as backend in general. All libraries used in our research are pip installable.

\section{Efficiency of explanation generation}
We also measure the efficiency for generating the explanations by the LIME and RAFT methods. The LIME method takes around 7 seconds to generate explanation for a text while \textsc{RAFT} models take around 1 second to generate an explanation. The RAFT models are 7 times more efficient than LIME.

For training the rationale predictor model, it took average of 1 hour/run. For the few shot experiments on the cross-domain dataset (50 datapoints) it takes 10-12 mins to train a single model. RGFS-SA and RGFS-CA takes 14\% and 18\% more time on average than the vanilla BERT models respectively.

\begin{table}[!t]
\centering
\scriptsize
\begin{tabular}{|l|p{6cm}|}
\hline
Target groups &  Categories\\
\hline
Race              & African, Arabs, Asians, Caucasian, Hispanic, Indian\\
Religion          & Buddhism, Christian, Hindu, Islam, Jewish, Non-religious\\
Gender            & Men, Women\\
Sexual Orient.    & Heterosexual, LGBTQ\\
Miscellaneous & Indigenous, Refugee/Immigrant, None, Others,Disability,Economic\\
\hline
\end{tabular}
\caption{\scriptsize{Target groups which occurred more than 20 times in the annotated dataset.}}
\label{tab:targets}
\end{table}

\begin{table}[!t]
\centering
\scriptsize
\begin{tabular}{lllllll}
\hline
\textbf{Dataset} & \textbf{Train} & \textbf{Val} & \textbf{Test} \\\hline
HX (S) & 15383 & 1922 & 1924 \\
FA  & 69859 & 9881 & 20059  \\
DA & 17347 & 2479  & 4957    \\
OD     & 9869  & 1396  & 2834    \\
BA   & 8963  & 1281  & 2561    \\
WH   & 7635   & 2192  & 1080   \\\hline
\end{tabular}
\caption{\scriptsize{This table shows the number of datapoints in train, validation and test data. The HateXplain dataset is divided into 8:1:1 ratio and other datasets are divided into 7:1:2 ratio into different splits.}}
\label{tab:dataset_stat}
\end{table}

\begin{table}[!t]
    \centering
    \scriptsize
    \begin{tabular}{cccc}
    \hline
    \textbf{Dataset} & \textbf{\#samples} & \textbf{Jaccard} &\textbf{random Jaccard}\\\hline
    FA   & 34 & 0.67  &0.34 \\
    DA   & 50 & 0.61  &0.33     \\
    OD   & 35  & 0.66 &0.27    \\
    BA   & 50  & 0.58 &0.32     \\
    WH   & 40   & 0.57 &0.26  \\\hline
    \end{tabular}
    \caption{\scriptsize{This table shows the number of samples annotated as abusive by both annotators out of the 50 samples per dataset, the Jaccard overlap between the annotated rationales (Jaccard) and random rationales (random Jaccard).}}
    \label{tab:annotated_jaccard}
\end{table}

\begin{table*}
\centering
\scriptsize
\begin{tabular}{ll}
\toprule
Model & Text \\
\midrule
Human annotation (OD dataset) &  user user user that expected if you placate the \hl{violent} \hl{leftists/} \hl{terrorists.} kavanaugh confirmation woke \\ \midrule
BERT &  user user user that expected if \hlr{you} placate the violent leftists / terrorists. \hlr{kavanaugh} \hlr{confirmation} \hlr{woke}\\
BERT-L-DOM & \hlr{user} \hlr{user} \hlr{user} that expected \hlr{if} you placate \hlr{the} \hlg{violent} \hlg{leftists/} terrorists. \hlr{kavanaugh} confirmation woke \\
RAFT-SA/CA & user user user that expected if you \hlr{placate} the \hlg{violent} \hlg{leftists/} \hlg{terrorists}. kavanaugh confirmation woke\\
\midrule
\midrule
Human annotation (BA dataset) &  user user user user a very high wall must be build to protect usa from \hl{bad} \hl{elements} \hl{of} \hl{illigal} 
\hl{refugees.} \\ \midrule
BERT &  \hlr{user} user user \hlr{user} \hlr{a} very high wall must be \hlr{build} to protect usa \hlr{from} bad elements \hlg{of} \hlg{illigal} refugees.\\
BERT-L-DOM &  user user \hlr{user} user \hlr{a} very high wall \hlr{must} be build to protect usa from \hlg{bad} elements \hlg{of} \hlg{illigal} \hlg{refugees.} \\
RAFT-SA/CA &   user user user user a very high wall must be build to protect \hlr{usa} \hlr{from} \hlg{bad} \hlg{elements} \hlg{of} \hlg{illigal} \hlg{refugees.}\\
\bottomrule
\end{tabular}
\caption{\scriptsize{Examples of rationales predicted by different models compared to human annotators. The first row corresponds to the annotation done by humans highlighted in \hl{yellow}. The \hlg{green highlight} represents the tokens which human annotators and the model found important. The \hlr{orange highlight} represents the tokens which the model found important, but the human annotators did not. \textsc{BERT-L-DOM} is the best cross domain model taken from Table \ref{tab:cross-domain-sim} corresponding to each target dataset.}}
\label{tab:compare explanation}
\end{table*}

\section{How do rationales help?}

In this section, we discuss further insights from our findings. We argue that human-like rationales play a very important role in learning subjective tasks like hate speech, sarcasm etc., as in the absence of such rationales, models can often focus on artefacts to get good performances. This is also evident from Table \ref{tab:compare explanation}, where the LIME based explanation focuses on artefact words present in the post like `user'\footnote{The anonymised version of mention.}, `if', `must', `build' etc. On the other hand, the rationales learnt using our BERT-RLT model are near perfect; this is also highlighted through the plausibility measurement in Table~\ref{tab:plausibility}. 

Furthermore, the rationale prediction is in the zero-shot setting, i.e., none of the target datasets contain labeled rationales that the model is fine-tuned upon. The performance will further improve if we can include few labelled rationale annotations~\cite{bhat2021self}. Since rationale annotation is more costly for the annotator/moderator, such zero-shot/few-shot rationale predictors can be very useful for reducing the overall workload of annotation. We would like to point out that similar to any machine learning algorithm, such rationale predictors can be erroneous. Appropriate feedback loops may be set to correct the model in those cases.

Ideally, we would like to have an explainable model which predicts the correct label along with the correct reasons. Current post-hoc explanations like LIME, although designed as faithful, may or may not provide correct/plausible reasons behind the prediction. Our rationale based attention framework outperforms LIME in terms of plausibility (noted in Table~\ref{tab:plausibility}) and performs comparably in terms of faithfulness metrics (as noted in the Table \ref{tab:faithfulness}). We believe this is a step in the right direction and future research in this direction can further develop better methods to add rationales.

\begin{table}[!t]
\centering
\scriptsize
\begin{tabular}{lllll}\hline
\multirow{2}{*}{\textbf{Model}}    & \multirow{2}{*}{\textbf{Data}}     & \multicolumn{3}{c}{\textbf{Plausibility}} \\\cline{3-5}
                          &                           & AUPRC    & token-F1  & IOU-F1    \\\hline
\textsc{BERT-L-DOM + LIME}& \multirow{4}{*}{DA}       & 0.77  & 0.52 & 0.21 \\
\textsc{BERT-L-DOM + SHAP}&                           & 0.45  & 0.37 & 0.14 \\
\textsc{RAFT}-CA/SA       &                           &\textbf{0.84}  & \textbf{0.58} & \textbf{0.26} \\\hline
\textsc{BERT-L-DOM +LIME} & \multirow{4}{*}{OD}       & 0.49  & 0.36 & 0.10\\
\textsc{BERT-L-DOM +SHAP} &                           & 0.37  & 0.32 & 0.07\\
\textsc{RAFT}-CA/SA       &                           & \textbf{0.68}  & \textbf{0.54} & \textbf{0.11} \\\hline
\textsc{BERT-L-DOM + LIME}& \multirow{4}{*}{BA}       & 0.63  & 0.43 & 0.19 \\
\textsc{BERT-L-DOM +SHAP} &                           & 0.46  & 0.34 & 0.14 \\
\textsc{RAFT}-CA/SA       &                           & \textbf{0.76}  & \textbf{0.55} & \textbf{0.23} \\\hline
\textsc{BERT-L-DOM + LIME}& \multirow{4}{*}{FA}       & 0.61  & 0.46 & 0.11  \\
\textsc{BERT-L-DOM + SHAP}&                           & 0.48  & 0.36 & 0.06  \\
\textsc{RAFT}-CA/SA       &                           & \textbf{0.67}  & \textbf{0.54} & \textbf{0.13} \\\hline
\textsc{BERT-L-DOM + LIME}& \multirow{4}{*}{WH}       & 0.57  & 0.42 & 0.01 \\
\textsc{BERT-L-DOM + SHAP} &                           & 0.50  & 0.35 & 0.01 \\
\textsc{RAFT}-CA/SA       &                           & \textbf{0.84}  & \textbf{0.63} & 0.01  \\\hline

\end{tabular}
\caption{\scriptsize{Average AUPRC, token-F1 and IOU-F1 scores for the rationales predicted by the models which were trained using different sets of 50 datapoints. For the models not utilising rationales in their architecture, LIME and SHAP are used to predict the rationales. \textsc{BERT-L-DOM} is the best cross-domain model for each dataset.}}
\label{tab:plausibility}
\end{table}

\begin{table}[!t]
\centering
\scriptsize
\begin{tabular}{llcc}\hline
\multirow{2}{*}{\textbf{Model}}   & \multirow{2}{*}{\textbf{Data}}     & \multicolumn{2}{l}{\textbf{Faithfulness}} \\\cline{3-4}
                                &                     & Suff.($\downarrow$)    & Comp.\\\hline
\textsc{BERT-L-DOM + LIME}       & \multirow{5}{*}{DA}& -0.03  & \textbf{0.67}  \\
\textsc{BERT-L-DOM + SHAP}       &                    & \textbf{-0.29}  & -0.14  \\
\textsc{RAFT}-CA          &                           & -0.06  & 0.11  \\
\textsc{RAFT}-SA          &                           & -0.08  & 0.25 \\\hline
\textsc{BERT-L-DOM + LIME}       & \multirow{5}{*}{OD}& -0.02  & 0.09 \\
\textsc{BERT-L-DOM + SHAP}       &                    & -0.10  & -0.09 \\
\textsc{RAFT}-CA          &                           & -0.04  & 0.04  \\
\textsc{RAFT}-SA          &                           & -0.08  & \textbf{0.29} \\\hline
\textsc{BERT-L-DOM + LIME}       & \multirow{5}{*}{BA}& -0.05  & \textbf{0.34} \\
\textsc{BERT-L-DOM + SHAP}       &                    & -0.38  & -0.38 \\
\textsc{RAFT}-CA          &                           & -0.04  & 0.06  \\
\textsc{RAFT}-SA          &                           & \textbf{-0.07}  & 0.19 \\\hline
\textsc{BERT-L-DOM + LIME}       & \multirow{5}{*}{FA}& -0.02  & \textbf{0.40} \\
\textsc{BERT-L-DOM + SHAP}       &                    & -0.09  & 0.09 \\
\textsc{RAFT}-CA          &                           & \textbf{-0.13}  & 0.09  \\
\textsc{RAFT}-SA          &                           & -0.11  & 0.26 \\\hline
\textsc{BERT-L-DOM + LIME} & \multirow{5}{*}{WH}       & -0.07  & \textbf{0.20} \\
\textsc{BERT-L-DOM + SHAP}       &                    & -0.09  & -0.09 \\
\textsc{RAFT}-CA          &                           & 0.01   & 0.03  \\
\textsc{RAFT}-SA          &                           & 0.06   & 0.06 \\\hline

\end{tabular}
\caption{\scriptsize{Average comprehensiveness scores (Comp) and the sufficiency scores (Suff) for the rationales predicted by the models which were trained using different sets of 50 datapoints. For the models not utilising rationales in their architecture, LIME and SHAP are used to predict the rationales. \textsc{BERT-L-DOM} is the best cross domain model corresponding to each dataset. For sufficiency scores, lower values are better.}}
\label{tab:faithfulness}
\end{table}

\end{document}